# Ensemble feature selection with data-driven thresholding for Alzheimer's disease biomarker discovery

Annette Spooner[1]*, Gelareh Mohammadi[1], Perminder S. Sachdev[2], Henry Brodaty[2], Arcot Sowmya[1]
for the Sydney Memory and Ageing Study and the Alzheimer's Disease Neuroimaging Initiative‡

[1] School of Computer Science and Engineering, UNSW Sydney, Sydney, Australia.
[2] Centre for Healthy Brain Ageing (CHeBA), School of Psychiatry, UNSW Sydney, Sydney, Australia.

*Corresponding author: Annette Spooner (a.spooner@unsw.edu.au)

## Abstract

Healthcare datasets present many challenges to both machine learning and statistics as their data are typically heterogeneous, censored, high-dimensional and have missing information. Feature selection is often used to identify the important features but can produce unstable results when applied to high-dimensional data, selecting a different set of features on each iteration.

The stability of feature selection can be improved with the use of feature selection ensembles, which aggregate the results of multiple base feature selectors. A threshold must be applied to the final aggregated feature set to separate the relevant features from the redundant ones. A fixed threshold, which is typically applied, offers no guarantee that the final set of selected features contains only relevant features. This work develops several data-driven thresholds to automatically identify the relevant features in an ensemble feature selector and evaluates their predictive accuracy and stability.

To demonstrate the applicability of these methods to clinical data, they are applied to data from two real-world Alzheimer's disease studies. Alzheimer's disease (AD) is a progressive neurodegenerative disease with no known cure, that begins at least 2-3 decades before overt symptoms appear, presenting an opportunity for researchers to identify early biomarkers that might identify patients at risk of developing AD. Features identified by applying these methods to both datasets reflect current findings in the AD literature.

## 1. Introduction

Healthcare datasets present many challenges to both machine learning and statistics alike. Their data are often heterogeneous, censored, high-dimensional and have missing information. In high-dimensional datasets, with a large number of features or variables and a small number of samples, typically only a small proportion of features may be relevant to the condition under investigation. Feature selection is generally used to reduce the dimension of the data, improve understanding of the problem and produce more interpretable models by identifying the factors that are important in understanding a disease [1]. A reliable model with fewer features can also lead to the development of more cost-effective procedures for identifying patients at risk of a disease.

However, applying feature selection to high-dimensional datasets often produces unstable results [2]. Stability, or reproducibility, of feature selection can be defined as the robustness of the selected features to perturbations in the data [3]. One of the key problems in modelling high-dimensional data, particularly where there are many redundant features, is the high variance of the models and feature selectors trained on this data. The same feature selection algorithm may select different subsets of features when run on different samples of the data, while achieving a similar level of predictive accuracy [4]. If clinicians are to have confidence in machine learning models developed on healthcare datasets, then the results of these models must be reproducible and therefore generalisable to new data.

Feature selection ensembles aggregate the results of multiple base feature selectors to improve stability and predictive accuracy [5]. One drawback of ensemble feature selectors, however, is that they do not provide a natural threshold to separate the relevant features from the irrelevant ones. A common method of setting this threshold is to choose a pre-determined fixed percentage of the number of features [6], but this offers no guarantee that the final set of selected features contains only relevant features. A data-driven threshold could overcome this problem and free the user from having to select and test different fixed thresholds [7].

Typically, simple univariate filters have been used as the base feature selectors in feature selection ensembles because they are computationally inexpensive, but they do not account for the interactions between features. More recently, feature selection ensembles have been created from multivariate models [8] [9] and these models provide more opportunities for developing data-driven thresholds.

The focus of this work is to develop and test several data-driven thresholding methods in a novel context, that of ensemble feature selection using multivariate base feature selectors. The first method uses the 75% quartile of the feature importance scores as a threshold. The second method uses kernel density estimation (KDE) to cluster the feature importance scores and exclude the irrelevant features based on these clusters. The final method uses permutations of random probes [10], which are random features that have no association with the target variable, to identify the relevant features. While these methods have been used in other contexts, to the best of our knowledge they have not previously been applied to ensemble feature selection.

To demonstrate the applicability of these methods to clinical data, they are applied to data from two quite different real-world Alzheimer's disease studies. Alzheimer's disease (AD) is a progressive neurodegenerative disease that results in declining cognitive function, such as memory, reasoning ability and executive function, and is ultimately fatal. Although the cause of this disease is not completely understood, the underlying pathological processes begin at least two to three decades before overt symptoms appear [11]. This presents an opportunity for researchers to determine early biomarkers that might help identify patients at risk of developing AD.

Ensemble feature selection has rarely been applied outside the context of classification. Data in healthcare, however, are often censored, meaning that the event of interest has not occurred during the study period, so the final outcome is unknown. Censored data are particularly common in AD studies as it is a slowly developing disease. The presence of censored data precludes the use of standard classification and regression techniques, but several machine learning algorithms have been adapted for survival analysis to handle censored data. It is these algorithms that we have chosen to investigate in this study, with the aim of expanding the use of ensemble feature selection to survival analysis. The methods developed here are applied to three real-world Alzheimer's disease (AD) datasets to identify biomarkers for AD.

The remainder of this paper is organised as follows. Section 2 reviews related work on ensemble feature selection and thresholding and aggregation of variable importance scores in other contexts. The experimental framework using data driven thresholds with ensemble feature selection is introduced in Section 3 and its results are presented in Section 4. Section 5 provides a discussion about the method and results and Section 6 concludes the work.

## 2. Related Work

### 2.1. Ensemble Feature Selection

Ensemble feature selectors apply a base feature selector to multiple subsamples of the training data, aggregate the results and apply a threshold to the resulting feature set. In the same way that bagging, boosting and stacking can improve the performance and reduce the variance of supervised learning methods, ensemble feature selection aims to improve the stability and predictive accuracy of the final subset of selected features [12]. The choice of the aggregation and thresholding methods are key elements that must be considered in the development of feature selection ensembles.

Ensemble feature selection was first proposed by Saeys et al. [6], who constructed homogeneous ensembles using simple filters applied to 40 bootstrap samples of the data and a simple linear sum of the feature rankings as aggregator. They compared the stability and performance of individual feature selection techniques with those of the ensembles and found that, in general, the ensemble techniques were more stable and had a similar predictive accuracy.

Other researchers have since conducted similar experiments, varying the feature selectors, the aggregators and the number of feature subsets in the ensemble[13] [14] [15] [16] [17] [9] [8] .

Most existing studies of feature selection ensembles use only simple filters as base methods, as these are the most computationally efficient methods to apply to high-dimensional data. The filters chosen most frequently include chi-squared, information gain and ReliefF [13] [18] [9]. A notable exception [8] examines three sparse feature selectors, which return a subset of important features – regularised regression, a tree-based gradient boosting machine and a deep neural network. Pes [9] also investigated a range of feature selection algorithms – univariate and multivariate, filters and embedded methods – in homogeneous ensembles in various application domains. But the use of multivariate feature selectors in ensemble feature selection is rare.

## 2.2 Thresholding of Feature Selection Ensembles

Much of the research on ensemble feature selection applies one or more fixed thresholds to the final feature selection in order to identify the most important features. In the case of gene rank aggregation, where the number of genes can run into the thousands or even tens of thousands, a threshold of 1% of the total number of features is common [13] [6] [16]. Various other values have been suggested, including $\log_2(n)$ where n is the total number of features [17], 5% [6], 10% and 20% [9] of the total number of features.

As the number of relevant features is not known a-priori, a fixed threshold could include some irrelevant features or alternatively reject some relevant ones. Data-driven thresholding can potentially overcome this problem and also free the user from having to select and test different fixed thresholds for each model [7].

Very little work has been reported on developing an automatic or data-driven thresholding method for ensemble feature selectors and this is an open area of research. Some early related works used the "biggest gap" between consecutive aggregated values as a point of threshold [19]. While intuitively this method has merit, in practice it may result in a very small or a very large final feature subset and is not always reliable.

Seijo-Pardo et al. [20] experimented with the use of three different data complexity measures to set automatic thresholds. However, these measures are only applicable to classification, and not to analysis of censored data.

Other researchers have investigated how to separate the important features from the redundant ones in individual feature selectors that return an importance score for each feature, such as random forests [21] [22] [23] [10]. Many of these methods use 'random probes' [21] [22] [23] to determine this boundary. A random probe is a random variable that has no association with the target variable and is typically created by randomly permuting the values of the existing features, thereby maintaining the same statistical distribution as the original features. Random probes are inserted as additional features into the data and the idea is that these features should be ranked last or at least as low as other irrelevant features. Features that are ranked below the probe can be discarded. The method of random probes is used in the popular Boruta feature selector [23]. Huynh-Thu et al. [10] were the first to suggest examining random probes in the context of stability analysis.

One-dimensional clustering of the feature importance scores is another method that may be useful in determining a threshold. The popular K-Means clustering method [24] has several limitations that make it unsuitable for this application, however. It requires the specification of the number of clusters a-priori, cannot detect non-spherical clusters and does not account for cluster density [25]. MAP-DP [25] is an alternative to K-means and was developed to overcome the limitations of K-means. It can handle clusters of different shapes and determines the number of clusters from the data, but it requires many other parameters to be specified, and so needs an in-depth knowledge of the algorithm.

Rodriguez and Liao [39] proposed a method of non-parametric clustering using kernel density estimation (KDE), that is able to find the correct number of clusters and detect non-spherical clusters. The cluster centres are defined as the local maxima in the density of data points. Once the centres have been identified, each point is assigned to the same cluster as its nearest neighbour of higher density. KDE overcomes the limitations of K-means clustering [24] and MAP-DP [25].

## 2.2. Aggregation of Feature Selection Ensembles

The problem of combining several ranked lists into a single final ranked list has been studied in fields as diverse as information retrieval, voting theory and bioinformatics, and various techniques have been proposed [18]. In statistics, this is known as the consensus ranking problem – "given m rankings of n objects, which ranking best represents the consensus opinion?" [26]. The *m* rankings may contain ties, be incomplete, and may be weighted.

Simple mathematical combinations, such as the mean, median or sum of the feature ranks or weights, are effective aggregation techniques and widely used [13] [6] [14] [16]. A count of the number of times each feature is selected by the base methods is also a commonly used method [27] [16]. Intuitively, if a feature is consistently given a high ranking in different data samples or by different methods, then it is likely to be important.

Wald et al. [18] carried out an extensive comparison of nine different rank aggregation techniques across twenty-six bioinformatics datasets and noted certain similarities between them. For example the Borda Count [28], a method often used in voting theory, is mathematically equivalent to the arithmetic mean and will rank features in the same order.

In some cases, aggregation and thresholding are combined. Kolde et al. [29] proposed an algorithm called Robust Rank Aggregation (RRA) for prioritised lists of genes, that assigns a significance score for each gene. It determines the probability that a randomly-generated rank list would have scored that feature more highly. The lower this probability, the more important the feature is. This method not only ranks the features but provides a statistically relevant threshold as well.

In the field of information retrieval, the number of items being ranked is usually very large and so it is not feasible to access every item in the database to calculate an aggregated score. More efficient techniques are needed for tasks such as the ranking of web search results or legal documents . The threshold algorithm (TA) is a simple yet elegant algorithm that allows early stopping and yields the top $k$ features, where $k$ must be chosen in advance [30]. Elements are accessed sequentially from the ranked lists i.e. the first element of each list is examined first, then the second element of each list and so on. At each sequential access a threshold equal to the sum of the scores in that access is set. For any items seen in the current access, random accesses are made to sample a set of scores for that element and an aggregate is calculated. A list of the highest scoring $k$ items seen so far is maintained and when all items in that set are greater than or equal to the threshold, the algorithm terminates. Therefore, it performs both aggregation and thresholding.

Another algorithm from the domain of information retrieval that can perform both aggregation and thresholding is the MedRank algorithm [31]. The MedRank algorithm also accesses the rankings sequentially. When an element has appeared in more than half of the ranked lists, it is output to the aggregated ranking. The algorithm can terminate early if only the top $k$ rankings are required.

This work proposes three different data-driven thresholding techniques for ensemble feature selection, adapted from other areas of research and applied in a novel context. These techniques are tested with two different real-world AD datasets and compared to a selection of fixed thresholds.

## 2.3. Stability Measures

Somol and Novovičová [32] studied measures of feature selection stability and noted some desirable properties. The measure should be bounded by 0 and 1 where a value of 1 should imply a high level of stability, whereas a value of 0 should imply a low level of stability. The measure should also be capable of evaluating the stability of feature sets of varying sizes.

Kuncheva's *consistency index* has been used by several authors investigating the stability of feature selection subsets [13] [16] [8], but it can only be applied to subsets of identical size. Lustgarten's *adjusted stability measure* [33] handles subsets of varying sizes, but does not fulfil the other desired properties. Somol and Novovičová's [32] *relative weighted consistency* of a set of feature subsets meets the desired properties and does not overemphasise low-frequency features.

## 3. Methods

### 3.1. Study Cohort

Experiments in this work were conducted on data from two real-world AD datasets - the Sydney Memory and Ageing Study [34] and the Alzheimer's Disease Neuroimaging Initiative [35]. The characteristics of both studies are summarised in Table 1 and a brief description of each is given in Sections 3.1.1 – 3.1.2. Full details can be found in the references provided.

The two datasets are quite different in terms of their study cohorts, data collected and depth of investigation and as such their features are not comparable. Instead, the aim in applying the methods developed here to these two datasets is to demonstrate their applicability to clinical data.

|  | MAS | ADNI-1 |
|---|---|---|
| **Study design** | Population based cohort study | Multisite longitudinal study |
| **Sample size (n)** | 873 | 819 |
| **Number of features (p)** | 140 | 216 |
| **Censoring rate** | 93% | 47% |
| **Intervals between waves** | 2 years | After 3, 6, 12, 18, 24, 36 and 48 months |
| **Age at baseline** | 70-90 years | 55 – 90 years |
| **Number of cases of AD** | 64 | 437 |

*Table 1. Study characteristics*

### 3.1.1. Sydney Memory and Ageing Study (MAS)

The Sydney Memory and Ageing Study (MAS) is a population-based cohort study aimed at examining the characteristics and prevalence of mild cognitive impairment and dementia. Full details of the study can be found [34]. The MAS data set contains a diverse collection of data including demographics, genetics, cognitive data, medical history, family history, medical examination, blood test results, psychological scores and functional data. Data that were used in forming a diagnosis of AD have not been used in the models developed here to predict AD.

The experiments reported in this paper used only the baseline data, collected in the first wave of MAS. Participants from a non-English-speaking background were excluded, leaving 873 participants from the original 1037. The event of interest in the survival analysis was a diagnosis of possible or probable Alzheimer's disease, over a period of 6 years, from wave 1 to wave 4 of the study. During this period 64 people developed Alzheimer's disease, indicating a censoring rate of 93%.

The Human Research Ethics Committees of the University of New South Wales and the South Eastern Sydney and Illawarra Area Health Service granted ethics approval for the MAS study and written consent was given by all participants and informants. The MAS study and this work were carried out in accordance with the MAS Governance guidelines, which are based on relevant University of New South Wales and National Health and Medical Research Council research and ethics policies.

### 3.1.2. Alzheimer's Disease Neuroimaging Initiative (ADNI)

The ADNI was launched in 2003 as a public-private partnership, led by Principal Investigator Michael W. Weiner, MD. The primary goal of ADNI has been to test whether serial magnetic resonance imaging (MRI), positron emission tomography (PET), other biological markers, and clinical and neuropsychological assessment can be combined to measure the progression of mild cognitive impairment (MCI) and early Alzheimer's disease (AD).

ADNI participants were aged 55 - 90 years at enrolment and were recruited from 57 sites in the United States and Canada. The ADNI data set contains data from a clinical evaluation, neuropsychological tests, genetic testing, lumbar puncture, and MRI and PET scans. Subjects who participated in ADNI phase 1 were selected for this study. The event of interest in the survival analysis was a diagnosis of probable AD, over the period of the ADNI1 study. A total of 200 participants with early AD were enrolled at the start of the study and a further 237 participants developed AD during the course of the study. Data that were used in forming a diagnosis of AD, have not been used in the models developed here to predict AD.

## 3.2. Experimental Framework

To prepare the data for the feature selection algorithms, missing data were imputed using the method of multiple imputation by chained equations in the R package *mice* [36]. Imputation was performed within the cross-validation loop.

Continuous features were normalised, by subtracting the mean and dividing by the standard deviation, and multiple values for the same measurement, e.g. blood pressure, were averaged. Levels of categorical features containing only a small number of samples were combined where possible. Further details of pre-processing steps can be found [37].

The R[38] package *mlr* (Machine Learning in R) [39] was used as a basis to carry out the experiments, while customised code was written to construct the ensembles. All of the ensembles were constructed within a 5-fold cross-validation framework, repeated 5 times. Random probes were generated for each subsample of the data. Experiments were performed on the computational cluster Katana, supported by Research Technology Services at UNSW Sydney [40].

### 3.3. Base Feature Selectors

The ensemble feature selectors were constructed from six different base feature selectors, each capable of selecting features from high-dimensional, heterogeneous, censored data. Four sparse methods, which return a subset of important features, and two filter methods, returning a score for each feature, were chosen. The four sparse methods were penalised regression for the Cox model (specifically the LASSO [41] and the ELASTIC-NET [42]), the Cox model with gradient boosting (GLMBOOST [43]) and the Cox model with likelihood-based boosting (COXBOOST [44]). The two filter methods were the maximally selected rank statistics random forests (RANGER [45]) and a univariate Cox filter (UNI). Each represents a different style of feature selection algorithm. The Cox filter was the only univariate method – the others are all multivariate feature selectors.

The absolute values of the coefficients of the features were used as feature importance scores for the sparse models - the LASSO, ELASTIC-NET, GLMBOOST and COXBOOST. These coefficients are meaningful importance scores because the data were normalised within the cross-validation loop prior to modelling. The other two models provide a feature importance score for each feature. For the RANGER, this was calculated using the method of permutation importance and for the univariate filter, the feature importance score was the value of the C-Index returned by a Cox Proportional Hazards model applied to each feature individually.

Further information about the functioning of these methods and the R packages used to implement them can be found [37].

Each of these feature selectors was first tested in its individual form. Within a framework of 5 repeats of 5-fold cross validation, the feature selector was applied to the training data to select relevant features. Several fixed thresholds (10%, 25%, 33% of the total number of features) were applied to the results of the filter methods, but as the sparse methods already select a subset of features, no further thresholding was applied to their results. A Ridge survival analysis model was trained and tested on the reduced dataset and the performance and stability of the individual models were compared to those of the ensemble models. The Ridge was chosen because of its superior performance in previous experiments [37].

### 3.4. Ensemble Construction

Homogeneous feature selection ensembles were constructed by applying the same feature selector to 50 bootstrapped samples of the training data, producing 50 subsets of features, as shown in Figure 1. An aggregator was used to combine these feature subsets into a single set and the resulting feature set was used as input to a machine learning model, in this case a survival analysis model, to assess its accuracy.

Five different aggregators were used to combine the feature subsets into a final feature set. Three of these aggregators also perform thresholding:

1. **Mean rank (MR):** The features in each subset were ranked according to their feature importance scores or weights and the mean of these ranks across all subsets was taken as the final aggregated score. The features were again ranked by this aggregated score.
2. **Mean weight (MW):** The weights or feature importance scores were averaged across all feature subsets and the features were ranked by the aggregated scores.
3. **Robust Rank Aggregation (RRA)** [29]: This method calculates a p-value, a statistically significant threshold, for each feature. Only features with a p-value of less than 0.05 were selected. This method performs both aggregation and thresholding, so no other thresholding techniques were applied.
4. **Threshold Algorithm (TA):** In each fold of the training data, the number of items to be returned, *k*, was set as the mean length of the 50 feature subsets generated by the ensemble. Elements were accessed sequentially from the ranked lists. At each sequential access a threshold equal to the sum of the scores in that access was set. A list of the highest scoring *k* items seen so far was maintained. When all items in that set were greater than or equal to the threshold, the algorithm terminated. This algorithm performs

both aggregation and thresholding, so no other thresholding techniques were applied
5. **Medrank Algorithm (MA):** In each fold of the training data, the number of features to be returned, *k*, was set as the mean length of the 50 feature subsets generated by the ensemble. Features were accessed sequentially from the ranked lists. When a feature appeared in more than 20% of the ranked lists, it was output to the aggregated list. (Here we used 20% rather than 50% as in the original algorithm because the signal is weak and very few features appear in more than half the lists). If the length of the aggregated list reached *k*, the algorithm terminated early.

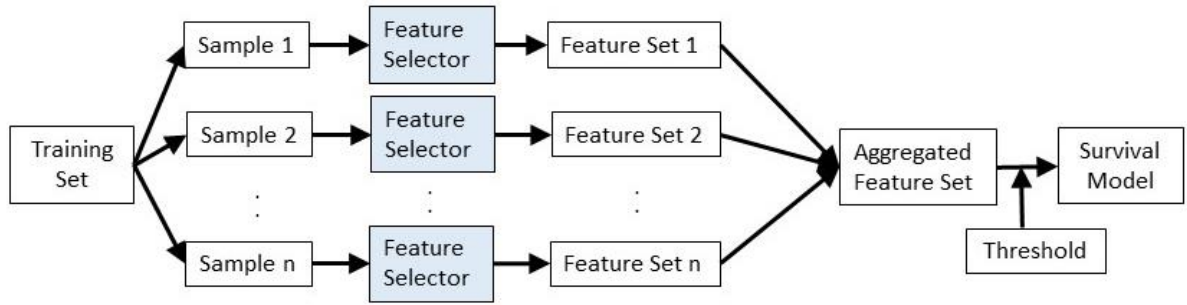

*Figure 1. A homogeneous feature selection ensemble. Sample1, Sample 2 … Sample n are randomly sampled subsets of the training data. The same feature selector is applied separately to each sample, generating n sets of selected features. An aggregator is applied to combine these feature sets into a single set, a threshold is applied and the resulting feature set is used as input to a survival model to assess its accuracy.*

A mixture of fixed and data driven thresholds were tested for comparison. Three different fixed thresholds were chosen - 10%, 25% and 33% of the total number of features, applied after aggregation.

Three different data-driven thresholds were also tested. The first method used the 75% quartile of the feature importance scores as a threshold. In preliminary testing the 25% and 50% quartiles were also examined, but the 75% quartile provided consistently better results, and so only that method has been included here.

The second method used kernel density estimation (KDE) to cluster the 1-dimensional importance scores, and this is the first time KDE has been used as a threshold in the context of ensemble feature selection. The cluster centres are defined as the local maxima in the density of data points. Each point is assigned to the same cluster as its nearest neighbour of higher density. In the case of 1-dimensional data, KDE can be plotted, as shown in Figure 2. The green dots show the local maxima or cluster centres. The red dots show the local minima, which are the boundaries between clusters. A key assumption of the method proposed here is that the majority of features are irrelevant, which is often the case in high-dimensional data. Then the maximum peak in the kernel density plot will be the cluster centre for the irrelevant features. Therefore, any features with an importance score higher than the upper boundary of that cluster (i.e. higher than the next local minimum), are the relevant features.

KDE determines the number of clusters from the data but still requires a bandwidth (also called a smoothing parameter) to be selected. Here the well-known rule of thumb, Silverman's rule, was used to select the bandwidth [46]. If the selected bandwidth was too large and produced only a single local maximum, and no local minima, a smaller value was generated by multiplying the original value by a factor of 0.75, and the density estimation was repeated. If a valid value was not found, no thresholding was performed.

The final method of data-driven thresholding tested used random probes to determine the boundary between the relevant and irrelevant features [21] [22] [23] [10]. A random probe is a random variable that has no association with the target variable and is typically created by randomly permuting the values of the existing features, thereby maintaining the same statistical distribution but breaking the correlation with the target variable. The values of the probes were randomly permuted on each of the 50 iterations. Any feature that was ranked below the rank of the highest random probe rank was considered irrelevant. This is the first time that random probes have has been used in the context of ensemble feature selection, although a similar idea was suggested by Huynh-Thu et al. [10].

Random probes were originally developed for use with continuous numeric data and feature selectors that return a score for each feature, so some adjustment was necessary to use them with heterogeneous data and sparse feature selectors. First, it is possible that a sparse feature selector may not select any of the random probes, meaning that none are given a feature importance score, and therefore a comparison cannot be made between the importance of the probes and the importance of the features. In this case all of the features selected by the sparse feature selector were considered relevant. Second, randomly permuting Boolean features (with values limited to TRUE or FALSE) and categorical features (with values limited to a small set of possible values) can leave some values in the same position, and so some of the random probes can achieve quite high importance scores, potentially eliminating truly important features. The random probes generated from Boolean features were excluded from the dataset for this reason. However, the random probes generated from categorical features were retained as the larger number of possible values ensured a more random permutation.

### 3.5. Performance Metrics

The prediction accuracy of the feature selection ensembles was assessed by the value of the Concordance Index (C-Index) achieved by a RIDGE survival model trained on the features selected by the ensemble. The Ridge model was chosen for its superior performance and stability in prior experiments [37]. The C-Index measures the proportion of pairs where the observation with the higher actual survival time has the higher probability of survival as predicted by the model [47]. The performance score for the ensemble was the mean of the performance scores over five repeats of 5-fold cross validation.

Each ensemble feature selector or individual method generated 25 final feature subsets from the 5 repeats of 5-fold cross validation. The stability of each ensemble was measured by applying Somol and Novovičová's *relative weighted consistency* [32] to these 25 feature subsets. This metric was chosen as it is capable of evaluating feature selectors that yield subsets of varying size.

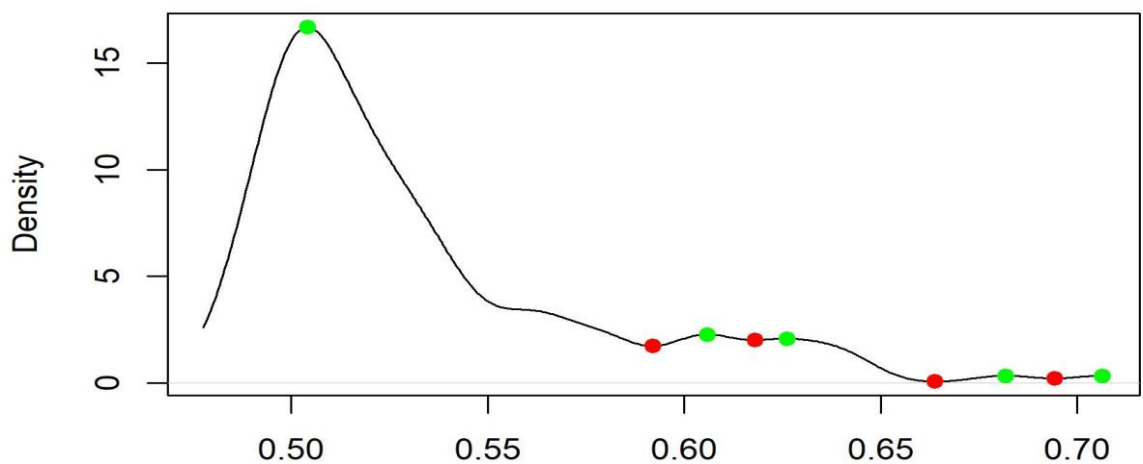

*Figure 2 - Example plot of kernel density estimate for one-dimensional clustering. The green dots show the local maxima, which are the cluster centres. The red dots show the local minima, which are the cluster boundaries. The maximum of the local maxima is the cluster centre for the irrelevant features.*

### 4. Results

The aim of this work was to develop and test a variety of data-driven thresholds for use with homogeneous feature selection ensembles, so as to free the user from having to select a fixed threshold. Three methods of data-driven thresholding were applied in a novel context and evaluated – the 75% quartile of the feature importance scores, KDE and the best random probe score. Three existing methods of thresholding, that combine thresholding with aggregation, were also tested – the RRA, and the Medrank and Threshold algorithms from the field of information retrieval. Three fixed thresholds were included for comparison. The thresholding methods were applied to homogeneous feature selection ensembles constructed from six different base feature selectors, capable of handling high-dimensional, right-censored data, and the stability and predictive accuracy of the ensembles were compared to their equivalent individual form. The features selected by these models were also examined as possible biomarkers for Alzheimer's disease.

Following the method of Song et al. [8], the predictive accuracy of the models was plotted against their stability in the graphs in Figure 3 for the MAS dataset, Figure 4 for the ADNI dataset. Values further to the right of the graph indicate more stable models and values higher on the graph indicate models with a higher predictive accuracy.

Note that the sparse feature selectors are designed to select a subset of the most useful features, therefore in their individual form no further thresholding is applied and there is only a single result for the individual form of the model (represented by a pale blue star). However, in the case of the filter methods, which return a score for each feature, a threshold must always be applied, even in the individual form, to select the most useful features. Therefore, the filter methods have multiple results for the individual form of the model, one for each fixed threshold and for KDE.

Song et al. [8] compared the overall performance of the models using the Euclidean distance from the origin in the plots of predictive accuracy vs stability. Using this method, the best ensemble model overall for the MAS dataset was RANGER with the RRA threshold, and the best ensemble model for the ADNI dataset was the UNI with the Threshold threshold. Both of these models use a data-driven threshold.

The graphs in Figures 3-4 show that most of the ensemble models are more stable than the individual form of the same model, with a value further to the right on the graph. In the MAS dataset, the greatest improvement in the Euclidean distance from the origin is once again seen in the RANGER model with the RRA threshold. In this case the ensemble shows an increase of 0.29 or 34% over the least stable individual model. In the ADNI dataset, the UNI model with the Threshold threshold shows the greatest improvement, with an increase of 0.32 or 32% over the least stable individual model. These improvements are mainly due to improved stability, which can be observed in the graphs in Figures 3-4.

The best threshold overall was determined by taking the average Euclidean distance from the origin for all models using each threshold. These results can be seen in the graph in Figure 5 and in Table 2 where the thresholds are listed in order of performance. In the MAS dataset the top four performing thresholds are data-driven thresholds, and these outperform the three fixed thresholds, while in the ADNI dataset the 10% fixed threshold is the top performing threshold, but the other two fixed thresholds are the worst performing.

The fixed thresholds show varying performance across the two datasets, showing that a fixed threshold must be tailored to the data at hand and carefully selected. In the MAS dataset, the performance of the three fixed thresholds are in the middle of the range, while in the ADNI dataset, which has a different number of features and samples, the fixed thresholds exhibit both the best and worst performance. It is clear that a fixed threshold must be carefully chosen to suit the dataset under investigation.

| MAS | | ADNI | |
| --- | --- | --- | --- |
| **Threshold** | 0.97 | **0.1** | 1.12 |
| **RRA** | 0.97 | **Best Probe** | 1.11 |
| **75% quantile** | 0.96 | **KDE** | 1.10 |
| **Medrank** | 0.94 | **RRA** | 1.10 |
| **0.1** | 0.94 | **Medrank** | 1.09 |
| **0.25** | 0.94 | **Threshold** | 1.08 |
| **0.33** | 0.94 | **75% quantile** | 1.07 |
| **KDE** | 0.87 | **0.25** | 1.06 |
| **Best Probe** | 0.87 | **0.33** | 1.05 |

*Table 2. Average Euclidean distance from the origin for each Threshold in the ADNI and MAS datasets, ordered from best to worst in each dataset, to show the relative performance of the methods.*

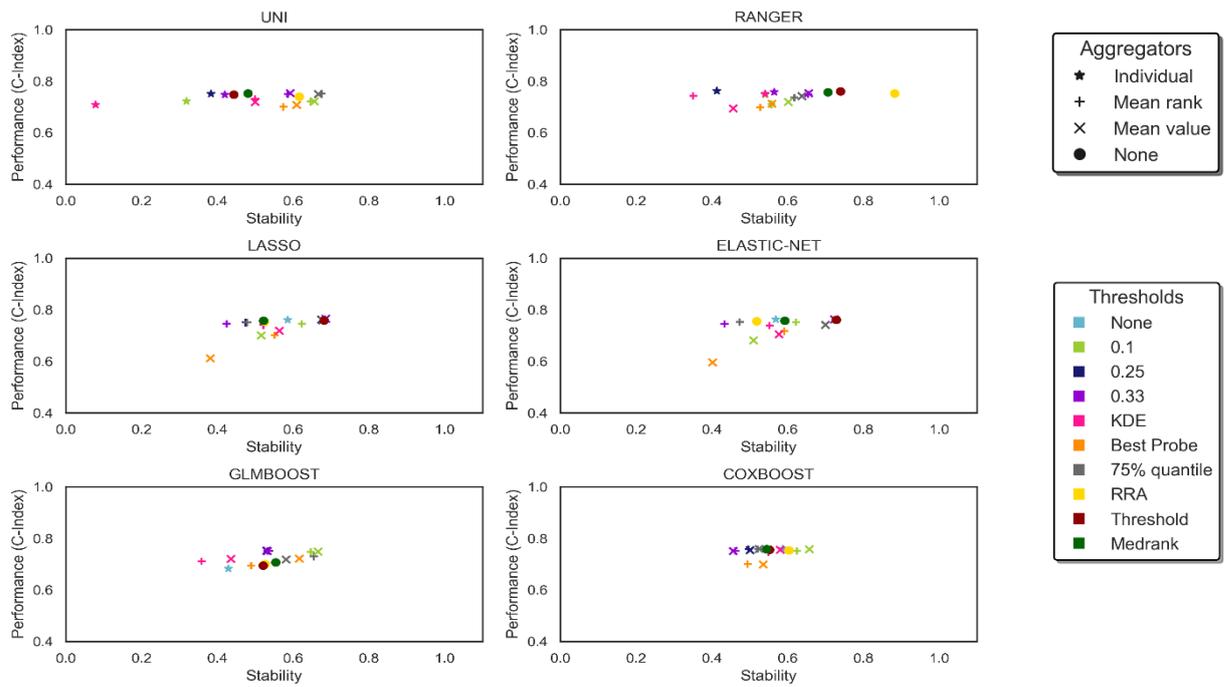

3. Experimental results from the MAS dataset. Each plot shows performance vs stability of one feature selector. The different shapes represent different aggregators, with a star shape representing the individual form, where the model is run only once and there is no aggregation of results. The different colours represent the different thresholds applied to the models.

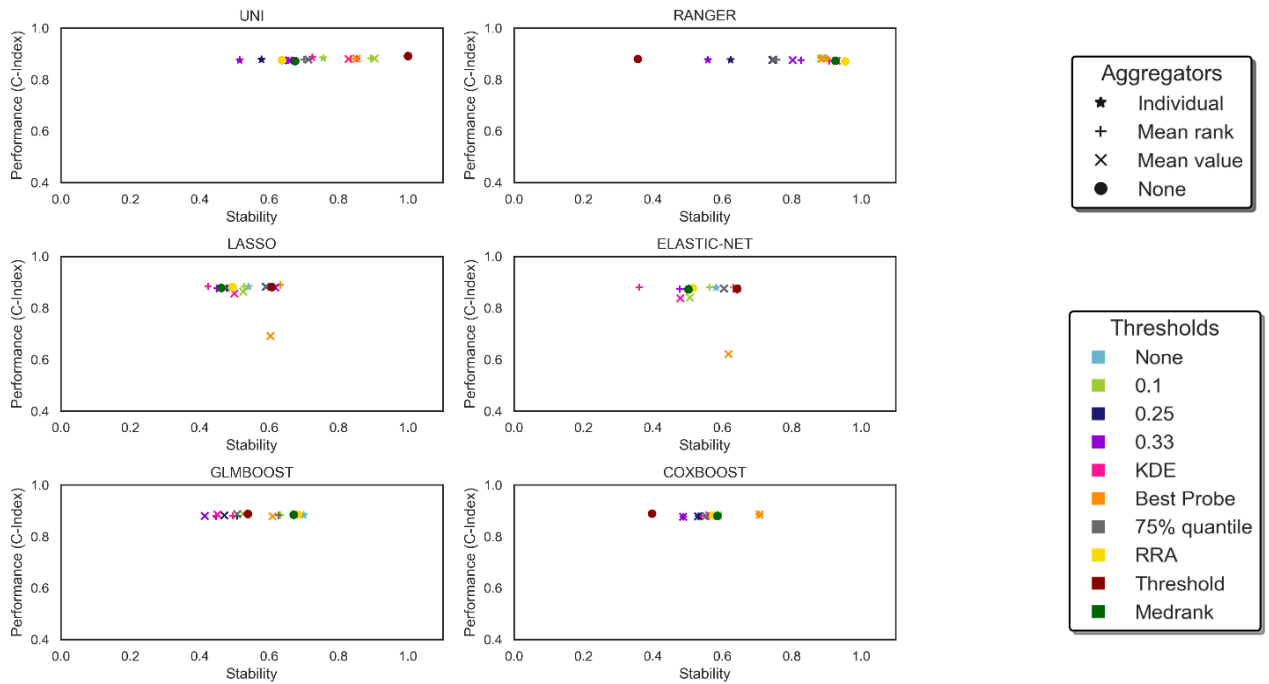

Figure 4. Experimental results from the ADNI dataset. Each plot shows performance vs stability of one feature selector. The different shapes represent different aggregators, with a star shape representing the individual form, where the model is run only once and there is no aggregation of results. The different colours represent the different thresholds applied to the models.

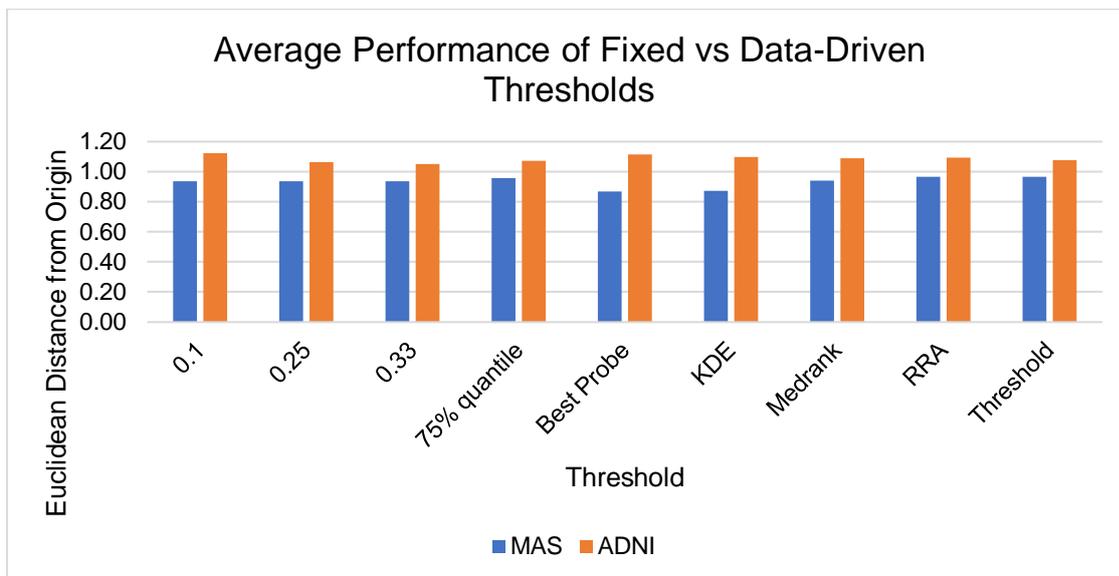

*Figure 5. Average Euclidean distance from the origin for each threshold for the MAS and ADNI dataset.*

## 4.1 Features Selected

The features selected by the ensemble feature selectors may be indicative biomarkers for AD. However, because of the stochastic nature of machine learning, even with improved stability each combination of feature selector, aggregator and threshold can still return a slightly different set of features. So, a decision must be made as to which is the optimal set of selected features.

Here we decided to identify the optimal features as those selected in at least 80% of the top 10 performing models. Features that are selected consistently by models that perform well are likely to be important features. The features selected in the MAS and ADNI datasets are described in Tables 3-4.

There are significant differences between the ADNI and OATS datasets. MAS is a population-based study which recruited non-demented individuals randomly from the community. ADNI recruited people with a diagnosis of AD or cognitively healthy individuals as volunteers. There are also differences in the data collected by the studies. ADNI had a strong focus on neuroimaging (structural MRI and PET) and biomarkers from cerebrospinal fluid (CSF) and plasma. In contrast, MRI imaging was performed on only a subset of MAS patients, PET imaging was not performed, no CSF was collected and fewer blood tests were performed in the baseline wave of MAS. MAS had a stronger focus on medical history and examination, psychological and neuro-psychological assessment and self-administered lifestyle questionnaires. Therefore, the features selected in the two datasets differ.

The AD literature supports the findings from both datasets. Age is known to be the most important predictor of dementia [48] . The epsilon4 allele of the APOE gene is known to increase the risk of late-onset AD [11]. Tests such as the Mini Mental State Exam (MMSE) [49], the General Practitioner Assessment of Cognition (GPCOG) [50] and the Informant Questionnaire on Cognitive Decline in the Elderly (IQCODE) [51] are designed to identify dementia. Evidence exists that atypical changes in driving behaviours may be early signs of mild cognitive impairment (MCI) and dementia [52] [53]. Some features, such as abnormal gait or posture can be related to increasing frailty, which is also associated with dementia [54]. A decline in the sense of smell [55] has been suggested as an early predictor of dementia and cardiovascular risk factors and obesity have also been linked to dementia [56]. Subjective cognitive complaints, from both participant and informant, are increasingly being recognised as predictors of progression to mild cognitive impairment and dementia [57]. The Neuropsychiatric Inventory can be used to detect behavioural changes in AD [58]. A recent review highlighted the link between rheumatoid arthritis and dementia [59] although other researchers have reported a negative link with AD rather than dementia generally [60]. Studies have found a link between low serum uric acid levels and AD [61]. Urinary tract infections are known to exacerbate dementia symptoms.

Other selected features are unique to the ADNI dataset. Many clinical studies have confirmed that the presence of tau and phosphorylated tau in CSF accurately detects Alzheimer's disease pathology [62] and amyloid-β (Aβ42) is also thought to be a biomarker for AD [63]. Increased concentrations of both CSF and plasma levels of the neurofilament light chain have been identified as potential biomarkers of AD [64] [65].

Patients with AD show an increase in CSF neurogranin concentration and this is not seen in other neuro-degenerative diseases [66]. Inflammation is thought to play a significant role in the pathophysiology of AD, and it has been shown that people with AD have a higher blood neutrophil-lymphocyte ratio (NLR), a marker of inflammation, than healthy controls [67]. Depression has been linked to AD and cognitive decline [68], as has a low level of education [56]. The assessment of instrumental activities of daily living (IADLs) is often used to determine the cognitive function of an individual [69].

| Feature Description - MAS |
| --- |
| Participant age at time of testing |
| Status of the epsilon4 allele of the APOE gene |
| Waist to hip ratio |
| Framingham cardio-vascular risk score |
| General Practitioner Assessment of Cognition score (GPCOG) |
| Mini Mental State Exam score (MMSE) |
| Informant Questionnaire on Cognitive Decline in the Elderly (IQCODE) |
| Informant subjective cognitive complaints – total score |
| Participant subjective cognitive complaints – total score |
| Composite variable encoding the number of major and minor at fault motor vehicle accidents in the past 18 months. |
| Normal or abnormal posture |
| Normal or abnormal gait |
| Urinary tract infection |
| Arthritis |
| Urate |
| Uric acid |

Table 3. Description of the features selected by the best models in MAS.

| Feature Description - ADNI |
| --- |
| Mini Mental State Exam score |
| Instrumental activities of daily living score |
| Status of the epsilon 4 allele of the APOE gene |
| 42-amino-acid-long beta amyloid peptide (abeta42) level in CSF |
| Plasma neurofilament light level |
| CSF neurofilament light level |
| CSF tau level |
| CSF phosphorylated tau level |
| Neuropsychiatric Inventory – total score |
| CSF neurogranin concentration |
| Geriatric depression scale score |
| Percentage of neutrophils |
| Percentage of lymphocytes |

Table 4. Description of the features selected by the best models in ADNI.

## 5. Discussion

The development of a data-driven threshold that can be applied to ensemble feature selectors for both censored and uncensored data is a distinct advantage. A fixed threshold that is appropriate for one data set may be quite inappropriate for another, especially when those datasets vary greatly in terms of the number and type of features they contain. The successful use of a fixed threshold involves testing of a wide range of fixed thresholds during the development of each model and can be quite time consuming.

The data-driven thresholds tested here performed well, being amongst the top performing thresholds in both datasets. The RRA threshold has the added advantage that it provides a p-value - a statistically relevant threshold, and so its use is highly recommended, particularly with clinical data.

The disadvantage of the KDE threshold is that it requires that the data contain a large number of irrelevant features to be successful. The KDE threshold performs better in the ADNI dataset than in the MAS dataset. This is because the ADNI dataset contains more features and so it is likely that it also has more irrelevant features.

The Best Probe threshold with the mean value aggregator did not perform well with the LASSO or ELASTICNET in either dataset. On investigation it was found that a random probe was often one of the most important features selected by the LASSO, thereby eliminating other truly important features. This could be the result of correlations in the data as the Lasso is known to select one feature at random from a group of correlated features. Further investigation is warranted to clarify this.

The Medrank and Threshold methods are primarily aggregators, and both output a fixed number of features, set in this case as the mean length of the ranked lists of features. Therefore, varying this number could affect their performance. Despite this, both perform well in these experiments, demonstrating the importance of an effective aggregation strategy.

As well as investigating different thresholds, a number of different feature selectors were examined in this work, including simple filters, penalised regression, random forests and boosted models. Although most previous works on ensemble feature selection have used only filters, it is clear from these results that multivariate feature selectors can also benefit by being used in an ensemble.

The ADNI dataset exhibited better performance and stability than the MAS dataset, which was to be expected. ADNI is an observational cohort study, where the number of patients with AD was carefully controlled at the start. Not only are there more cases of AD in ADNI, giving it a much lower censoring rate than MAS, but participants with AD at baseline were accepted into the study. In contrast, MAS is a population-based cohort study, where participants were excluded if they had AD at baseline, and so the number of AD cases cannot be controlled and there are fewer cases of AD than in ADNI.

In future it would be of interest to train the models on one AD dataset and apply them to another AD dataset with a comparable cohort and set of features, but the aim here was to demonstrate the applicability of the methods to different datasets, albeit in the same area. The methods were applied to a third dataset, which had few events and fewer samples than the other datasets. The results were mixed and have not been included here but they demonstrates the need for an adequate amount of data for these methods to succeed.

Future work could investigate the use of the Knockoffs method for variable selection [70], rather than random probes. Like random probes, knockoffs are random variables that have no association with the target variable. Whereas random probes maintain the same statistical distribution as the original variables, knockoffs maintain the same correlation structure between the original variables and the target variable.

Future work could also apply the techniques developed here to heterogeneous ensembles of feature selectors, where a different feature selector is applied to each sample of the training data. Using different feature selectors would introduce more diversity into the ensemble and so may produce enhanced results.

## 6.   Conclusion

This study has demonstrated the use and validity of data-driven thresholding methods applied to ensemble feature selectors, to provide more stable, and therefore more reproducible, selections of features than individual feature selectors, without loss of performance. Several methods of data-driven thresholding have been used in a novel context and have been shown to perform well.  The use of a data-driven threshold eliminates the need to choose a threshold a-priori and can select a more meaningful set of features. A reliable and compact set of features can produce more interpretable models by identifying the factors that are important in understanding a disease [1] and can also lead to the development of more cost-effective procedures for identifying patients at risk of a disease.

A number of multivariate feature selectors were tested for use in feature selection ensembles, in contrast to the univariate filters that are typically employed in this context. Issues arising from the application of the data-driven thresholds to heterogeneous data and multivariate feature selectors, particularly those that select a subset of features rather than returning a score for each feature, were overcome, allowing these methods to be used in feature selection ensembles.

The ability to produce more stable selections of features means that clinicians can have more confidence in the results produced by machine learning models. Features that are predictive of Alzheimer's disease have been selected from the models developed here and these are in keeping with findings in the AD literature.

**Acknowledgements**

The authors acknowledge the contribution of the MAS research team and administrative assistants to this article, for their advice and collection and management of data.

Funding: The MAS study was supported by a National Health and Medical Research Council of Australia Program Grant (ID 350833). We thank the MAS study participants for their time and generosity in contributing to this research.

‡ Data from the Alzheimer's Disease Neuroimaging Initiative (ADNI) database (adni.loni.usc.edu) were used in preparation of this article. As such, the investigators within the ADNI contributed to the design and implementation of ADNI and/or provided data but did not participate in analysis or writing of this report. A complete listing of ADNI investigators can be found at: http://adni.loni.usc.edu/wp-content/uploads/how_to_apply/ADNI_Acknowledgement_List.pdf

Funding: Data collection and sharing was funded by the ADNI (National Institutes of Health Grant U01 AG024904) and DOD ADNI (Department of Defense award number W81XWH-12-2-0012). ADNI is funded by the National Institute on Aging, the National Institute of Biomedical Imaging and Bioengineering, and through generous contributions from the following: AbbVie, Alzheimer's Association; Alzheimer's Drug Discovery Foundation; Araclon Biotech; BioClinica, Inc.; Biogen; Bristol-Myers Squibb Company; CereSpir, Inc.; Cogstate; Eisai Inc.; Elan Pharmaceuticals, Inc.; Eli Lilly and Company; EuroImmun; F. Hoffmann-La Roche Ltd and its affiliated company Genentech, Inc.; Fujirebio; GE Healthcare; IXICO Ltd.; Janssen Alzheimer Immunotherapy Research & Development, LLC.; Johnson & Johnson Pharmaceutical Research & Development LLC.; Lumosity; Lundbeck; Merck & Co., Inc.; Meso Scale Diagnostics, LLC.; NeuroRx Research;

Neurotrack Technologies; Novartis Pharmaceuticals Corporation; Pfizer Inc.; Piramal Imaging; Servier; Takeda Pharmaceutical Company; and Transition Therapeutics. The Canadian Institutes of Health Research is providing funds to support ADNI clinical sites in Canada. Private sector contributions are facilitated by the Foundation for the National Institutes of Health (www.fnih.org). The grantee organization is the Northern California Institute for Research and Education, and the study is coordinated by the Alzheimer's Therapeutic Research Institute at the University of Southern California. ADNI data are disseminated by the Laboratory for Neuro Imaging at the University of Southern California.

Finally, the authors thank Dr Jordan Raykov, of Aston University, Birmingham, UK, for his help in implementing the MAP-DP clustering algorithm in R.

**Author Contributions**

A.Sp. prepared the data, designed and ran the machine learning experiments, wrote the custom code and wrote the paper. A.So. and G.M. provided expert guidance and reviewed the manuscript. P.S. and H.B. provided data and had intellectual input into revising the manuscript.

**Corresponding Author**
Annette Spooner, UNSW Sydney, a.spooner@unsw.edu.au

**Competing Interests**
The authors declare that they have no competing interests.